\documentclass[10pt,journal,compsoc]{IEEEtran}
\ifCLASSOPTIONcompsoc
  \usepackage[nocompress]{cite}
\else
  \usepackage{cite}
\fi
\ifCLASSINFOpdf
  \usepackage[pdftex]{graphicx}
  \DeclareGraphicsExtensions{.pdf}
\else
\fi
\usepackage{amsmath}
\usepackage{amssymb}
\usepackage{url}

\usepackage{multirow}
\usepackage{booktabs}
\usepackage{amsfonts}
\usepackage{amsthm}

\begin{document}
%
\title{Second-Order Pooling for Graph Neural Networks}
%
%
%
%

\author{Zhengyang~Wang
    and~Shuiwang~Ji,~\IEEEmembership{Senior~Member,~IEEE}
    \IEEEcompsocitemizethanks{\IEEEcompsocthanksitem Zhengyang Wang and Shuiwang Ji are with the Department of Computer Science and Engineering, Texas A\&M University, College Station, TX, 77843.\protect\\
E-mail: sji@tamu.edu}
\thanks{Manuscript received October, 2019.}}

%
%

\markboth{IEEE Transactions on Pattern Analysis and Machine Intelligence,~Vol.~xx, No.~x, October~2019}%
{Wang \MakeLowercase{\textit{et al.}}: Second-Order Pooling for Graph Neural Networks}
%



\IEEEtitleabstractindextext{%
\begin{abstract}
Graph neural networks have achieved great success in learning node
representations for graph tasks such as node classification and link
prediction. Graph representation learning requires graph pooling to
obtain graph representations from node representations. It is
challenging to develop graph pooling methods due to the variable
sizes and isomorphic structures of graphs. In this work, we propose
to use second-order pooling as graph pooling, which naturally solves
the above challenges. In addition, compared to existing graph
pooling methods, second-order pooling is able to use information
from all nodes and collect second-order statistics, making it more
powerful. We show that direct use of second-order pooling with graph
neural networks leads to practical problems. To overcome these
problems, we propose two novel global graph pooling methods
based on second-order pooling; namely, bilinear mapping and attentional
second-order pooling. In addition, we extend attentional second-order
pooling to hierarchical graph pooling for more flexible use in GNNs.
We perform thorough experiments on graph classification
tasks to demonstrate the effectiveness and superiority of our
proposed methods. Experimental results show that our methods improve
the performance significantly and consistently.
\end{abstract}

}

\maketitle

\IEEEdisplaynontitleabstractindextext

%
\IEEEpeerreviewmaketitle

\IEEEraisesectionheading{\section{Introduction}\label{sec:intro}}

In recent years, deep learning has been widely explored on graph structured data, such as chemical compounds, protein structures, financial networks, and social networks~\cite{yanardag2015deep,zhang2018graph,wu2019comprehensive}. Remarkable success has been achieved by generalizing deep neural networks from grid-like data to graphs~\cite{fan2019graph,Gao:2019:CRF:3292500.3330888,ma2019disentangled,Wang:2019:HGA:3308558.3313562}, resulting in the development of various graph neural networks~(GNNs), like graph convolutional network~(GCN)~\cite{kipf2017semi}, GraphSAGE~\cite{hamilton2017inductive}, graph attention network~(GAT)~\cite{velivckovic2018graph}, jumping knowledge network~(JK)~\cite{xu2018representation}, and graph isomorphism networks~(GINs)~\cite{xu2019powerful}. They are able to learn representations for each node in graphs and have set new performance records on tasks like node classification and link prediction~\cite{schutt2017schnet}. In order to extend the success to graph representation learning, graph pooling is required, which takes node representations of a graph as inputs and outputs the corresponding graph representation.

While pooling is common in deep learning on grid-like data, it is challenging to develop graph pooling approaches due to the special properties of graphs. First, the number of nodes varies in different graphs, while the graph representations are usually required to have the same fixed size to fit into other machine learning models. Therefore, graph pooling should be capable of handling the variable number of node representations as inputs and producing fixed-sized graph representations. Second, unlike images and texts where we can order pixels and words according to the spatial structural information, there is no inherent ordering relationship among nodes in graphs. Indeed, we can set pseudo indices for nodes in a graph. However, an isomorphism of the graph may change the order of the indices. As isomorphic graphs should have the same graph representation, it is required for graph pooling to create the same output by taking node representations in any order as inputs.

Some previous studies employ simple methods such as averaging and
summation as graph
pooling~\cite{duvenaud2015convolutional,defferrard2016convolutional,xu2019powerful}.
However, averaging and summation ignore the feature correlation information,
hampering the overall model performance~\cite{zhang2018end}. Other
studies have proposed advanced graph pooling methods, including
\textsc{DiffPool}~\cite{ying2018hierarchical},
\textsc{SortPool}~\cite{zhang2018end},
\textsc{TopKPool}~\cite{gao2018graph},
\textsc{SAGPool}~\cite{lee2019self},
and \textsc{EigenPool}~\cite{Ma:2019:GCN:3292500.3330982}.
\textsc{DiffPool} maps nodes to a pre-defined number of clusters but is hard to train.
\textsc{EigenPool} involves the computation of eigenvectors, which is slow and expensive.
\textsc{SortPool}, \textsc{SAGPool} and \textsc{TopKPool} rely on the top-$K$ sorting
to select a fixed number~($K$) of nodes and order them, during which
the information from unselected nodes is discarded. It is worth
noting that all the existing graph pooling methods only collect
first-order statistics~\cite{boureau2010theoretical}.

In this work, we propose to use second-order pooling as graph
pooling. Compared to existing graph pooling methods, second-order
pooling naturally solves the challenges of graph pooling and is more
powerful with its ability of using information from all nodes and
collecting second-order statistics. We analyze the practical
problems in directly using second-order pooling with GNNs. To
address the problems, we propose two novel and effective global graph
pooling approaches based on second-order pooling; namely, bilinear
mapping and attentional second-order pooling. In addition, we extend
attentional second-order pooling to hierarchical graph pooling for more
flexible use in GNNs. We perform thorough
experiments on ten graph classification benchmark datasets. The
experimental results show that our methods improve the performance
significantly and consistently.

\section{Related Work}

In this section, we review two categories of existing graph pooling methods in Section~\ref{sec:gp}. Then in Section~\ref{sec:so}, we introduce what second-order statistics are, as well as their applications in both transitional machine learning and deep learning. In addition, we discuss the motivation of using second-order statistics in graph representation learning.

\subsection{Graph Pooling: Global versus Hierarchical}\label{sec:gp}

Existing graph pooling methods can be divided into two categories according to their roles in graph neural networks~(GNNs) for graph representation learning. One is global graph pooling, also known as graph readout operation~\cite{xu2019powerful,lee2019self}. The other is hierarchical graph pooling, which is used to build hierarchical GNNs. We explain the details of the two categories and provide examples. In addition, we discuss advantages and disadvantages of the two categories.

Global graph pooling is typically used to connect embedded graphs outputted by GNN layers with classifiers for graph classification. Given a graph, GNN layers produce node representations, where each node is embedded as a vector. Global graph pooling is applied after GNN layers to process node representations into a single vector as the graph representation. A classifier takes the graph representation and performs graph classification. The ``global'' here refers to the fact that the output of global graph pooling encodes the entire graph. Global graph pooling is usually used only once in GNNs for graph representation learning. We call such GNNs as flat GNNs, in contrast to hierarchical GNNs. The most common global graph pooling methods include averaging and summation~\cite{duvenaud2015convolutional,defferrard2016convolutional,xu2019powerful}.

Hierarchical graph pooling is more similar to pooling in computer vision tasks~\cite{boureau2010theoretical}. The output of hierarchical graph pooling is a pseudo graph with fewer nodes than the input graph. It is used to build hierarchical GNNs, where hierarchical graph pooling is used several times between GNN layers to gradually decrease the number of nodes. The most representative hierarchical graph pooling methods are \textsc{DiffPool}~\cite{ying2018hierarchical}, \textsc{SortPool}~\cite{zhang2018end}, \textsc{TopKPool}~\cite{gao2018graph}, \textsc{SAGPool}~\cite{lee2019self}, and \textsc{EigenPool}~\cite{Ma:2019:GCN:3292500.3330982}. A straightforward way to use hierarchical graph pooling for graph representation learning is to reduce the number of nodes to one. Then the resulted single vector is treated as the graph representation. Besides, there are two other ways to generate a single vector from the pseudo graph outputted by hierarchical graph pooling. One is introduced in \textsc{SAGPool}~\cite{lee2019self}, where global and hierarchical graph pooling are combined. After each hierarchical graph pooling, global graph pooling with an independent classifier is employed. The final prediction is an average of all classifiers. On the other hand, \textsc{SortPool}~\cite{zhang2018end} directly applies convolutional neural networks~(CNNs) to reduce the number of nodes to one. In particular, it takes advantage of a property of the pseudo graph outputted by hierarchical graph pooling. That is, the pseudo graph is a graph with a fixed number of nodes and there is an inherent ordering relationship among nodes determined by the trainable parameters in the hierarchical graph pooling. Therefore, common deep learning methods like convolutions can be directly used. In fact, we can simply concatenate node presentations following the inherent order as the graph representation.

Given this property, most hierarchical graph pooling methods can be flexibly used as global graph pooling, with the three ways introduced above. For example, \textsc{SortPool}~\cite{zhang2018end} is used to build flat GNNs and applied only once after all GNN layers. While the idea of learning hierarchical graph representations makes sense, hierarchical GNNs do not consistently outperform flat GNNs~\cite{lee2019self}. In addition, with advanced techniques like jumping knowledge networks~(JK-Net)~\cite{xu2018representation} to address the over-smoothing problem of GNN layers~\cite{chen2019measuring}, flat GNNs can go deeper and achieve better performance than hierarchical GNNs~\cite{xu2019powerful}.

In this work, we first focus on global graph pooling as second-order pooling naturally fits this category. Later, we extend one of our proposed graph pooling methods to hierarchical graph pooling in Section~\ref{sec:multihead}.

\subsection{Second-Order Statistics}\label{sec:so}

In statistics, the $k$-order statistics refer to functions which use the $k$-th power of samples. Concretely, consider $n$ samples $(x_1, x_2, \ldots, x_n)$. The first and second moments, \emph{i.e.}, the mean $\mu=\frac{1}{n}\sum_i x_i$ and variance $\sigma^2=\frac{1}{n}\sum_i (x_i - \mu)^2$, are examples of first and second-order statistics, respectively. If each sample is a vector, the covariance matrix is an example of second-order statistics. In terms of graph pooling, it is easy to see that existing methods are based on first-order statistics~\cite{boureau2010theoretical}.

Second-order statistics have been widely explored in various computer vision tasks, such as face recognition, image segmentation, and object detection. In terms of traditional machine learning, the scale-invariant feature transform (SIFT) algorithm~\cite{lowe1999object} utilizes second-order statistics of pixel values to describe local features in images and has become one of the most popular image descriptors. Tuzel et` al.\cite{tuzel2006region,tuzel2008pedestrian} use covariance matrices of low-level features with boosting for detection and classification. The Fisher encoding~\cite{perronnin2010improving} applies second-order statistics for recognition as well. Carreira et al.~\cite{carreira2012semantic} employs second-order pooling for semantic segmentation. With the recent advances of deep learning, second-order pooling is also used in CNNs for fine-grained visual recognition~\cite{lin2015bilinear} and visual question answering~\cite{gao2016compact,fukui2016multimodal,wang2018learning}.

Many studies motivates the use of second-order statistics as taking advantage of the Riemannian geometry of the space of symmetric positive definite matrices~\cite{arsigny2007geometric,tuzel2008pedestrian,carreira2012semantic}. In these studies, certain regularizations are cast to guarantee that the applied second-order statistics are symmetric positive definite~\cite{acharya2018covariance,wang2019deep}. Other work relates second-order statistics to orderless texture descriptors for images~\cite{perronnin2010improving,lin2015bilinear}.

In this work, we propose to incorporate second-order statistics in graph representation learning. Our motivations lie in three aspects. First, second-order pooling naturally fits the goal and requirements of graph pooling, as discussed in Sections~\ref{sec:method_pool} and~\ref{sec:method_sop}. Second, second-order pooling is able to capture the correlations among features, as well as topology information in graph representation learning, as demonstrated in Section~\ref{sec:method_sop}. Third, our proposed graph pooling methods based on second-order pooling are related to covariance pooling~\cite{tuzel2006region,tuzel2008pedestrian,acharya2018covariance,wang2019deep} and attentional pooling~\cite{girdhar2017attentional} used in computer vision tasks, as pointed out in Section~\ref{sec:method_discuss}. In addition, we show that both covariance pooling and attentional pooling have certain limitations when employed in graph representation learning, and our proposed methods appropriately address them.

\section{Second-Order Pooling for Graphs}\label{sec:method}

In this section, we introduce our proposed second-order pooling methods for graph representation learning. First, we formally define the aim and requirements of graph pooling in Section~\ref{sec:method_pool}. Then we propose to use second-order pooling as graph pooling, analyze its advantages, and point out practical problems when directly using it with GNNs in Section~\ref{sec:method_sop}. In order to address the problems, we propose two novel second-order pooling methods for graphs in Sections~\ref{sec:method_1} and~\ref{sec:method_2}, respectively. Afterwards, we discuss why our proposed methods are more suitable as graph pooling compared to two similar pooling methods in image tasks in Section~\ref{sec:method_discuss}. Finally, while both methods focus on global graph pooling, we extend second-order pooling to hierarchical graph pooling in Section~\ref{sec:multihead}.

\subsection{Properties of Graph Pooling}\label{sec:method_pool}

Consider a graph $G = (A, X)$ represented by its adjacency matrix $A \in \{0,1\}^{n \times n}$ and node feature matrix $X \in \mathbb{R}^{n \times d}$, where $n$ is the number of nodes in $G$ and $d$ is the dimension of node features. The node features may come from node labels or node degrees. Graph neural networks~(GNNs) are known to be powerful in learning good node representation matrix $H$ from $A$ and $X$:
\begin{equation}
H = [h_1, h_2, \ldots, h_n]^T = \text{GNN}(A, X) \in \mathbb{R}^{n \times f},
\end{equation}
where rows of $H$, $h_i \in \mathbb{R}^f, i = 1, 2, \ldots, n$, are representations of $n$ nodes, and $f$ depends on the architecture of GNNs. The task that we focus on in this work is to obtain a graph representation vector $h_G$ from $H$, which is then fed into a classifier to perform graph classification:
\begin{equation}\label{eqn:p}
h_G = g([A], H) \in \mathbb{R}^{c},
\end{equation}
where $g(\cdot)$ is the graph pooling function and $c$ is the dimension of $h_G$. Here, $[A]$ means that the information from $A$ can be optionally used in graph pooling. For simplicity, we omit it in the following discussion.

Note that $g(\cdot)$ must satisfy two requirements to serve as graph pooling. First, $g(\cdot)$ should be able to take $H$ with variable number of rows as the inputs and produce fixed-sized outputs. Specifically, different graphs may have different number of nodes, which means that $n$ is a variable. On the other hand, $c$ is supposed to be fixed to fit into the following classifier.

Second, $g(\cdot)$ should output the same $h_G$ when the order of
rows of $H$ changes. This permutation invariance property is
necessary to handle isomorphic graphs. To be concrete, if two graph
$G_1 = (A_1, X_1)$ and $G_2 = (A_2, X_2)$ are isomorphic, GNNs will
output the same multiset of node
representations~\cite{zhang2018end,xu2019powerful}. That is, there
exists a permutation matrix $P \in \{0,1\}^{n \times n}$ such that
$H_1 = PH_2$, for $H_1 = \text{GNN}(A_1, X_1)$ and $H_2 =
\text{GNN}(A_2, X_2)$. However, the graph representation computed by
$g(\cdot)$ should be the same, \emph{i.e.}, $g(H_1) = g(H_2)$ if
$H_1 = PH_2$.

\subsection{Second-Order Pooling}\label{sec:method_sop}

In this work, we propose to employ second-order pooling~\cite{carreira2012semantic}, also known as bilinear pooling~\cite{lin2015bilinear}, as graph pooling. We show that second-order pooling naturally satisfies the two requirements above.

We start by introducing the definition of second-order pooling.

\textbf{Definition.} Given $H = [h_1, h_2, \ldots, h_n]^T \in \mathbb{R}^{n \times f}$, second-order pooling~(\textsc{SOPool}) is defined as
\begin{equation}\label{eqn:sop_def}
\textsc{SOPool}(H) = \sum_{i=1}^{n} h_ih_i^T = H^TH \in \mathbb{R}^{f \times f}.
\end{equation}

In terms of graph pooling, we can view $\textsc{SOPool}(H)$ as an
$f^2$-dimensional graph representation vector by simply flattening
the matrix. Another way to transform the matrix into a vector is
discussed in Section~\ref{sec:method_2}. Note that, as long as
$\textsc{SOPool}$ meets the two requirements, the way to
transform the matrix into a vector does not affect its eligibility
as graph pooling.

\begin{figure*}
    \centering
    \includegraphics[width=\textwidth]{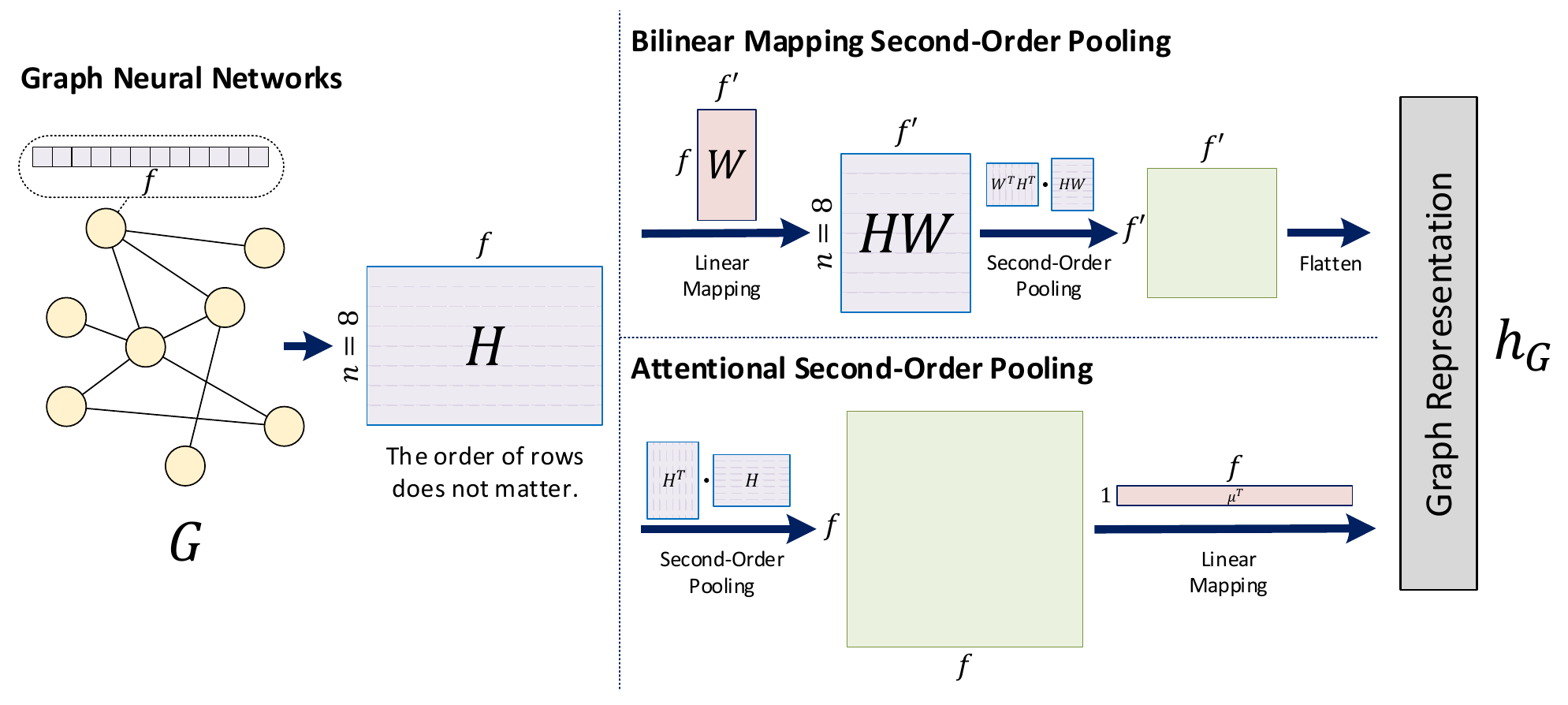}
    \caption{Illustrations of our proposed graph pooling methods: bilinear mapping second-order pooling~(\textsc{SOPool}$_{bimap}$) in Section~\ref{sec:method_1} and attentional second-order pooling~(\textsc{SOPool}$_{attn}$) in Section~\ref{sec:method_2}. This is an example for a graph $G$ with $n=8$ nodes. GNNs can learn representations for each node and graph pooling processes node representations into a graph representation vector $h_G$.}
    \label{fig:methods}
\end{figure*}

Now let us check the two requirements.

\textbf{Proposition 1.} $\textsc{SOPool}$ always outputs an $f \times f$ matrix for $H \in \mathbb{R}^{n \times f}$, regardless of the value of $n$.
\begin{proof}
    The result is obvious since the dimension of $H^TH$ does not depend
    on $n$.
\end{proof}

\textbf{Proposition 2.} $\textsc{SOPool}$ is invariant to permutation so that it
outputs the same matrix when the order of rows of $H$ changes.
\begin{proof} Consider $H_1 = PH_2$, where $P$ is a permutation matrix. Note that we have $P^TP = I$ for any permutation matrix. Therefore, it is easy to derive
    \begin{eqnarray}
    \textsc{SOPool}(H_1) &=& H_1^TH_1 \nonumber \\
    &=& (PH_2)^T(PH_2) \nonumber \\
    &=& H_2^TP^TPH_2 \nonumber \\
    &=& H_2^TH_2 = \textsc{SOPool}(H_2).
    \end{eqnarray}
    This completes the proof.
\end{proof}

In addition to satisfying the requirements of graph pooling, $\textsc{SOPool}$ is capable of capturing second-order statistics, which are much more discriminative than first-order statistics computed by most other graph pooling methods~\cite{carreira2012semantic,lin2015bilinear,gao2016compact}. In detail, the advantages can be seen from two aspects. On one hand, we can tell from $\textsc{SOPool}(H) = \sum_{i=1}^{n} h_ih_i^T$ that, for each node representation $h_i$, the features interact with each other, enabling the correlations among features to be captured. On the other hand, topology information is encoded as well. Specifically, we view $H \in \mathbb{R}^{n \times f}$ as $H = [l_1, l_2, \ldots, l_f]$, where $l_j \in \mathbb{R}^n$, $j=1,2,\ldots,f$. The vector $l_j$ encodes the spatial distribution of the $j$-th feature in the graph. Based on this view, $\textsc{SOPool}(H) = H^TH$ is able to capture the topology information.

However, we point out that the direct application of second-order pooling in GNNs leads to practical problems. The direct way to use second-order pooling as graph pooling is represented as
\begin{equation}
h_G = \textsc{Flatten}(\textsc{SOPool}(\text{GNN}(A, X))) \in \mathbb{R}^{f^2}.
\end{equation}
That is, we apply $\textsc{SOPool}$ on $H =
\text{GNN}(A, X)$ and flatten the output matrix into an
$f^2$-dimensional graph representation vector. However, it causes an
explosion in the number of training parameters in the following
classifier when $f$ is large, making the learning process harder to
converge and easier to overfit. While each layer in a GNN usually
has outputs with a small number of hidden units (e.g. 16, 32, 64),
it has been pointed out that graph representation learning benefits
from using information from outputs of all layers, obtaining better
performance and generalization ability~\cite{xu2018representation}.
It is usually achieved by concatenating outputs across all layers in
a GNN~\cite{zhang2018end,xu2019powerful}. In this case, $H$ has a
large final $f$, making direct use of second-order pooling infeasible. For example, if
a GNN has 5 layers and each layer's outputs have 32 hidden units,
$f$ becomes $32 \times 5 = 160$. Suppose $h_G$ is sent into a 1-layer fully-connected
classifier for $c$ graph categories in a graph classification task.
It results in $160^2c = 25,600c$ training parameters, which is excessive.
We omit the bias term for simplicity.

\subsection{Bilinear Mapping Second-Order Pooling}\label{sec:method_1}

To address the above problem, a straightforward solution is to reduce $f$ in $H$ before $\textsc{SOPool}(H)$. Based on this, our first proposed graph pooling method, called bilinear mapping second-order pooling~(\textsc{SOPool}$_{bimap}$), employs a linear mapping on $H$ to perform dimensionality reduction. Specifically, it is defined as
\begin{eqnarray}
\textsc{SOPool}_{bimap}(H) &=& \textsc{SOPool}(HW) \nonumber \\
&=& W^TH^THW \in \mathbb{R}^{f' \times f'},
\end{eqnarray}
where $f' < f$ and $W \in \mathbb{R}^{f \times f'}$ is a trainable
matrix representing a linear mapping. Afterwards, we follow
the same process to flatten the matrix and obtain an
$f'^2$-dimensional graph representation vector:
\begin{equation}
h_G = \textsc{Flatten}(\textsc{SOPool}_{bimap}(\text{GNN}(A, X))) \in \mathbb{R}^{f'^2}.
\end{equation}
Figure~\ref{fig:methods} provides an illustration of the above process. By selecting an appropriate $f'$, the bilinear mapping second-order pooling does not suffer from the excessive number of training parameters. Taking the example above, if we set $f' = 32$, the total number of parameters in \textsc{SOPool}$_{bimap}$ and a following 1-layer fully-connected classifier is $32 \times 160 + 32^2c = 5,120 + 1,024c$, which is much smaller than $25,600c$.

\subsection{Attentional Second-Order Pooling}\label{sec:method_2}

\begin{figure*}
    \centering
    \includegraphics[width=\textwidth]{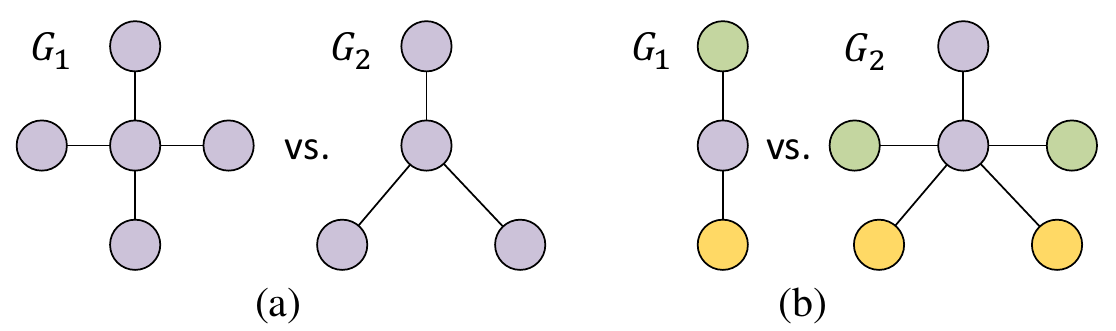}
    \caption{Examples of graphs that pooling methods discussed in Section~\ref{sec:method_discuss} fail to distinguish, \emph{i.e.}, producing the same graph representation for different graphs $G_1$ and $G_2$. The same color denotes the same node representation. (a) Covariance pooling~(\textsc{CovPool}) and attentional pooling~(\textsc{AttnPool}) both fail. \textsc{CovPool} fails because subtracting the mean results in $h_{G_1} = h_{G_2} = \boldsymbol{0}$. \textsc{AttnPool} computes the mean of node representations, leading to $h_{G_1} = h_{G_2}$ as well. (b) \textsc{AttnPool} fails in this example with the same $\mu$.}
    \label{fig:example}
\end{figure*}

Our second proposed graph pooling method tackles with the problem by exploring another way to transform the matrix computed by $\textsc{SOPool}$ into the graph representation vector, instead of simply flattening. Similarly, we use a linear mapping to perform the transformation, defined as
\begin{equation}
h_G = \textsc{SOPool}(\text{GNN}(A, X)) \cdot \mu \in \mathbb{R}^{f},
\end{equation}
where $\mu \in \mathbb{R}^{f}$ is a trainable vector. It is interesting to note that $h_G = H^TH\mu$, which is similar to the sentence attention in~\cite{yang2016hierarchical}. To be concrete, consider a word embedding matrix $E = [e_1, e_2, \ldots, e_l]^T \in \mathbb{R}^{l \times d_w}$ for a sentence, where $l$ is the number of words and $d_w$ is the dimension of word embeddings. The sentence attention is defined as
\begin{eqnarray}
\alpha_i &=& \frac{\exp(e_i^T\mu_s)}{\sum_{j=1}^{l}\exp(e_j^T\mu_s)}, i = 1, 2, \ldots, l \label{eqn:softmax} \\
s &=& \sum_{i=1}^{l}\alpha_ie_i,
\end{eqnarray}
where $\mu_s \in \mathbb{R}^{d_w}$ is a trainable vector and $s$ is the resulted sentence embedding. Note that Eqn.~(\ref{eqn:softmax}) is the \textsc{Softmax} function and serves as a normalization function~\cite{vaswani2017attention}. Rewriting the sentence attention into matrix form, we have $s = E^T\textsc{Softmax}(E\mu_s)$. The only difference between the computation of $h_G$ and that of $s$ is the normalization function. Therefore, we name our second proposed graph pooling method as attentional second-order pooling~(\textsc{SOPool}$_{attn}$), defined as
\begin{equation}\label{eqn:attn}
\textsc{SOPool}_{attn}(H) = H^TH\mu \in \mathbb{R}^{f},
\end{equation}
where $\mu \in \mathbb{R}^{f}$ is a trainable vector. It is illustrated in Figure~\ref{fig:methods}. We take the same example above to show that \textsc{SOPool}$_{attn}$ reduces the number of training parameters. The total number of parameters in \textsc{SOPool}$_{attn}$ and a following 1-layer fully-connected classifier is just $160 + 160c$, significantly reducing the amount of parameters compared to $25,600c$.

\subsection{Relationships to Covariance Pooling and Attentional Pooling}\label{sec:method_discuss}

The experimental results in Section~\ref{sec:exp} show that both our proposed graph pooling methods achieve better performance significantly and consistently than previous studies. However, we note that, there are pooling methods in image tasks that have similar computation processes to our proposed methods, although they have not been developed based on second-order pooling. In this section, we point out the key differences between these methods and ours and show why they matter in graph representation learning.

Note that images are usually processed by deep neural networks into feature maps $I \in \mathbb{R}^{h \times w \times c}$, where $h$, $w$, $c$ are the height, width, and number of feature maps, respectively. Following~\cite{girdhar2017attentional,acharya2018covariance,wang2019deep}, we reshape $I$ into the matrix $H \in \mathbb{R}^{n \times f}$, where $n = hw$ and $f = c$ so that different pooling methods can be compared directly.

\textbf{Covariance pooling.} Covariance pooling~(\textsc{CovPool})~\cite{tuzel2006region,tuzel2008pedestrian,acharya2018covariance,wang2019deep} has been widely explored in image tasks, such as image categorization, facial expression recognition, and texture classification. Recently, it has also been explored in GNNs~\cite{verma2018graph}. The definition is
\begin{equation}\label{eqn:cov_def}
\textsc{CovPool}(H) = (H-\boldsymbol{1}\bar{H})^T(H-\boldsymbol{1}\bar{H}) \in \mathbb{R}^{f \times f},
\end{equation}
where $\boldsymbol{1}$ is the $n$-dimensional all-one vector and $\bar{H} \in \mathbb{R}^{1 \times f}$ is the mean of rows of $H$. It differs from $\textsc{SOPool}$ defined in Eqn.~(\ref{eqn:sop_def}) only in whether to subtract the mean. However, subtracting the mean makes $\textsc{CovPool}$ less powerful in terms of distinguishing graphs with repeating node embeddings~\cite{xu2019powerful}, which may cause the performance loss. Figure~\ref{fig:example}(a) gives an example of this problem.

\textbf{Attentional pooling.} Attentional pooling~(\textsc{AttnPool})~\cite{girdhar2017attentional} has been used in action recognition. As shown in Section~\ref{sec:method_2}, it is also used in text classification~\cite{yang2016hierarchical}, defined as
\begin{equation}\label{eqn:attn_def}
\textsc{AttnPool}(H) = H^T\textsc{Softmax}(H\mu) \in \mathbb{R}^{f},
\end{equation}
where $\mu \in \mathbb{R}^{f}$ is a trainable vector. It differs from \textsc{SOPool}$_{attn}$ only in the \textsc{Softmax} function. We show that the \textsc{Softmax} function leads to similar problems as other normalization functions, such as mean and max-pooling~\cite{xu2019powerful}. Figure~\ref{fig:example} provides examples in which $\textsc{AttnPool}$ does not work.

To conclude, our methods derived from second-order pooling are more suitable as graph pooling. We compare these pooling methods through experiments in Section~\ref{sec:exp_pool}. The results show that $\textsc{CovPool}$ and $\textsc{AttnPool}$ suffer from significant performance loss on some datasets.

\subsection{Multi-Head Attentional Second-Order Pooling}\label{sec:multihead}

The proposed \textsc{SOPool}$_{bimap}$ and \textsc{SOPool}$_{attn}$ belong to the global graph pooling category. As discussed in Section~\ref{sec:gp}, they are used in flat GNNs after all GNN layers and output the graph representation for the classifier. While flat GNNs outperform hierarchical GNNs in most benchmark datasets~\cite{xu2019powerful}, developing hierarchical graph pooling is still desired, especially for large graphs~\cite{ying2018hierarchical,Ma:2019:GCN:3292500.3330982,lee2019self}. Therefore, we explore a hierarchical graph pooling method based on second-order pooling.

Unlike global graph pooling, hierarchical graph pooling outputs multiple vectors corresponding to node representations in the pooled graph. In addition, hierarchical graph pooling has to update the adjacency matrix to indicate how nodes are connected in the pooled graph. To be specific, given the adjacency matrix $A \in \mathbb{R}^{n \times n}$ and node representation matrix $H \in \mathbb{R}^{n \times f}$, a hierarchical graph pooling function $g_h(\cdot)$ can be written as
\begin{equation}\label{eqn:hp}
A', H' = g_h(A, H),
\end{equation}
where $A' \in \mathbb{R}^{k \times k}$ and $H' \in \mathbb{R}^{k \times f}$. Here, $k$ is a hyperparameter determining the number of nodes in the pooled graph. Note that Eqn.~(\ref{eqn:hp}) does not conflict with Eqn.~(\ref{eqn:p}), as we can always transform $H'$ into a vector $h_G$, as discussed in Section~\ref{sec:gp}.

We note that the proposed \textsc{SOPool}$_{attn}$ in Section~\ref{sec:method_2} is closely related to the attention mechanism and can be easily extended to a hierarchical graph pooling method based on the multi-head technique in the attention mechanism~\cite{vaswani2017attention,velivckovic2018graph}. The multi-head technique means that multiple independent attentions are performed on the same inputs. Then the outputs of multiple attentions are then concatenated together. Based on this insight, we propose multi-head attentional second-order pooling~(\textsc{SOPool}$_{m\_attn}$), defined as
\begin{equation}\label{eqn:mattn1}
H' = \textsc{SOPool}_{m\_attn}(H) = UH^TH \in \mathbb{R}^{k \times f},
\end{equation}
where $U \in \mathbb{R}^{k \times f}$ is a trainable matrix. To illustrate its relationship to the multi-head technique, we can equivalently write it as
\begin{equation}\label{eqn:mattn2}
\textsc{SOPool}_{m\_attn}(H) = [H^TH\mu_1, \ldots, H^TH\mu_k]^T,
\end{equation}
where we decompose $U$ in Eqn.~(\ref{eqn:mattn1}) as $U=[\mu_1, \mu_2, \ldots, \mu_k]^T$. The relationship can be easily seen by comparing Eqn.~(\ref{eqn:mattn2}) with Eqn.~(\ref{eqn:attn}).

The multi-head technique enables \textsc{SOPool}$_{m\_attn}$ to output the node representation matrix for the pooled graph. We now describe how to update the adjacency matrix. In particular, we employ a contribution matrix $C$ in updating the adjacency matrix. The contribution matrix is a $k \times n$ matrix, whose entries indicate how nodes in the input graph contribute to nodes in the pooled graph. In \textsc{SOPool}$_{m\_attn}$, we can simply let $C = UH^T \in \mathbb{R}^{k \times n}$. With the contribution matrix $C$, the corresponding adjacency matrix $A'$ of the pooled graph can be computed as
\begin{equation}\label{eqn:mattn_c}
A' = CAC^T \in \mathbb{R}^{k \times k}.
\end{equation}

The proposed \textsc{SOPool}$_{m\_attn}$ is closely related to \textsc{DiffPool}~\cite{ying2018hierarchical}. The contribution matrix $C$ corresponds to the assignment matrix in \textsc{DiffPool}. However, \textsc{DiffPool} applied GNN layers with normalization on $H$ to obtain $C$, preventing the explicit use of second-order statistics. In the experiments, we evaluate \textsc{SOPool}$_{m\_attn}$ as both global and hierarchical graph pooling methods, in flat and hierarchical GNNs, respectively.

\section{Experiments}\label{sec:exp}

We conduct thorough experiments on graph classification tasks to show the effectiveness of our proposed graph pooling methods, namely bilinear mapping second-order pooling~(\textsc{SOPool}$_{bimap}$), attentional second-order pooling~(\textsc{SOPool}$_{attn}$), and multi-head attentional second-order pooling~(\textsc{SOPool}$_{m\_attn}$). Section~\ref{sec:exp_setup} introduces the datasets, baselines, and experimental setups for reproduction. The following sections aim at evaluating our proposed methods in different aspects, by answering the questions below:
\begin{itemize}
	\item Can GNNs with our proposed methods achieve improved performance in graph classification tasks? Section~\ref{sec:exp_baseline} provides the comparison results between our methods and existing methods in graph classification tasks.
	\item Do our proposed methods outperform existing global graph pooling methods with the same flat GNN architecture? The ablation studies in Section~\ref{sec:exp_pool} compare different graph pooling methods with the same GNN, eliminating the influences of different GNNs. In particular, we use hierarchical graph pooling methods as global graph pooling methods in this experiment, including \textsc{SOPool}$_{m\_attn}$.
	\item Is the improvement brought by our proposed method consistent with various GNN architectures? Section~\ref{sec:exp_encoder} shows the performance of the proposed \textsc{SOPool}$_{bimap}$ and \textsc{SOPool}$_{attn}$ with different GNNs.
	\item Is \textsc{SOPool}$_{m\_attn}$ effective as hierarchical graph pooling methods? We compare \textsc{SOPool}$_{m\_attn}$ with other hierarchical graph pooling methods in the same hierarchical GNN architecture in Section~\ref{sec:exp_hierarchical}.
\end{itemize}

\subsection{Experimental Setup}\label{sec:exp_setup}

\textbf{Reproducibility.} The code used in our experiments is available at \url{https://github.com/divelab/sopool}. Details of datasets and parameter settings are described below.

\textbf{Datasets.} We use ten graph classification datasets from~\cite{yanardag2015deep}, including five bioinformatics datasets (MUTAG, PTC, PROTEINS, NCI1, DD) and five social network datasets (COLLAB, IMDB-BINARY, IMDB-MULTI, REDDIT-BINARY, REDDIT-MULTI5K). Note that only bioinformatics datasets come with node labels. Below are the detailed descriptions of datasets:

\begin{itemize}
    \item MUTAG is a bioinformatics dataset of 188 graphs representing nitro compounds. Each node is associated with one of 7 discrete node labels. The task is to classify each graph by determining whether the compound is mutagenic aromatic or heteroaromatic~\cite{debnath1991structure}.
    \item PTC~\cite{toivonen2003statistical} is a bioinformatics dataset of 344 graphs representing chemical compounds. Each node comes with one of 19 discrete node labels. The task is to predict the rodent carcinogenicity for each graph.
    \item PROTEINS~\cite{borgwardt2005protein} is a bioinformatics dataset of 1,113 graph structures of proteins. Nodes in the graphs refer to secondary structure elements (SSEs) and have discrete node labels indicating whether they represent a helix, sheet or turn. And edges mean that two nodes are neighbors along the amino-acid sequence or in space. The task is to predict the protein function for each graph.
    \item NCI1~\cite{wale2008comparison} is a bioinformatics dataset of 4,110 graphs representing chemical compounds. It contains data published by the National Cancer Institute (NCI). Each node is assigned with one of 37 discrete node labels. The graph classification label is decided by NCI anti-cancer screens for ability to suppress or inhibit the growth of a panel of human tumor cell lines.
    \item COLLAB is a scientific collaboration dataset of 5,000 graphs corresponding to ego-networks generated using the method in~\cite{shrivastava2014new}. The dataset is derived from 3 public collaboration datasets~\cite{leskovec2005graphs}. Each ego-network contains different researchers from each field and is labeled by the corresponding field. The three fields are High Energy Physics, Condensed Matter Physics, and Astro Physics.
    \item IMDB-BINARY is a movie collaboration dataset of 1,000 graphs representing ego-networks for actors/actresses. The dataset is derived from collaboration graphs on Action and Romance genres. In each graph, nodes represent actors/actresses and edges simply mean they collaborate the same movie. The graphs are labeled by the corresponding genre and the task is to identify the genre for each graph.
    \item IMDB-MULTI is multi-class version of IMDB-BINARY. It contains 1,500 ego-networks and has three extra genres, namely, Comedy, Romance and Sci-Fi.
    \item REDDIT-BINARY is a dataset of 2,000 graphs where each graph represents an online discussion thread. Nodes in a graph correspond to users appearing in the corresponding discussion thread and an edge means that one user responded to another. Datasets are crawled from top submissions under four popular subreddits, namely, IAmA, AskReddit, TrollXChromosomes, atheism. Among them, AmA and AskReddit are question/answer-based subreddits while TrollXChromosomes and atheism are discussion-based subreddits, forming two classes to be classified.
    \item REDDIT-MULTI5K is a similar dataset as REDDIT-BINARY, which contains 5,000 graphs. The difference lies in that REDDIT-MULTI5K crawled data from five different subreddits, namely, worldnews, videos, AdviceAnimals, aww and mildlyinteresting. And the task is to identify the subreddit of each graph instead of determining the type of subreddits.
    \item DD~\cite{dobson2003distinguishing} is a bioinformatics dataset of 1,178 graph structures of proteins. Nodes in the graphs represent amino acids. And edges connect nodes that are less than 6 $\mathring{A}$ngstroms apart. The task is a two-way classification task between enzymes and non-enzymes. DD is only used in Section~\ref{sec:exp_hierarchical}. The average number of nodes in DD is 284.3.
\end{itemize}

\begin{table*}
    \caption{Comparison results between our proposed methods and baselines described in Section~\ref{sec:exp_setup}. We report the accuracies of these baselines provided in \cite{zhang2018end,verma2018graph,xu2019powerful,ying2018hierarchical,Ma:2019:GCN:3292500.3330982}. The best models are highlighted with boldface. If a kernel-based baseline performs the best than all GNN-based models, we highlight the best GNN-based model with boldface and the best kernel-based baseline with boldface and asterisk.}
    \label{table:baseline}
    \centering
    \resizebox{1.0\textwidth}{!}{
        \begin{tabular}{clccccccccc}
            \toprule
            \multirow{5}{*}{\rotatebox[origin=c]{90}{Datasets}}
            && MUTAG & PTC & PROTEINS & NCI1 & COLLAB & IMDB-B & IMDB-M & RDT-B & RDT-M5K\\
            &\# graphs & 188 & 344 & 1113 & 4110 & 5000 & 1000 & 1500 & 2000 & 5000 \\
            &\# classes & 2 & 2 & 2 & 2 & 3 & 2 & 3 & 2 & 5 \\
            &\# nodes (max) & 28 & 109 & 620 & 111 & 492 & 136 & 89 & 3783 & 3783 \\
            &\# nodes (avg.) & 18.0 & 25.6 & 39.1 & 29.9 & 74.5 & 19.8 & 13.0 & 429.6 & 508.5 \\
            \midrule
            \multirow{5}{*}{\rotatebox[origin=c]{90}{Kernel}}
            &GK [2009] & 81.4$\pm$1.7 & 57.3$\pm$1.4 & 71.7$\pm$0.6 & 62.3$\pm$0.3 & 72.8$\pm$0.3 & 65.9$\pm$1.0 & 43.9$\pm$0.4 & 77.3$\pm$0.2 & 41.0$\pm$0.2 \\
            &RW [2010] & 79.2$\pm$2.1 & 57.9$\pm$1.3 & 74.2$\pm$0.4 & $>$1 day & - & - & - & - & - \\
            &WL [2011] & 90.4$\pm$5.7 & 59.9$\pm$4.3 & 75.0$\pm$3.1 & \textbf{86.0}$\pm$\textbf{1.8$^*$} & 78.9$\pm$1.9 & 73.8$\pm$3.9 & 50.9$\pm$3.8 & 81.0$\pm$3.1 & 52.5$\pm$2.1 \\
            &DGK [2015] & - & 60.1$\pm$2.6 & 75.7$\pm$0.5 & 80.3$\pm$0.5 & 73.1$\pm$0.3 & 67.0$\pm$0.6 & 44.6$\pm$0.5 & 78.0$\pm$0.4 & 41.3$\pm$0.2 \\
            &AWE [2018] & 87.9$\pm$9.8 & - & - & - & 73.9$\pm$1.9 & 74.5$\pm$5.9 & 51.5$\pm$3.6 & 87.9$\pm$2.5 & 54.7$\pm$2.9 \\
            \midrule
            \multirow{7}{*}{\rotatebox[origin=c]{90}{GNN}}
            &DCNN [2016] & 67.0 & 56.6 & 61.3 & 56.6 & 52.1 & 49.1 & 33.5 & - & - \\
            &\textsc{PatchScan} [2016] & 92.6$\pm$4.2 & 60.0$\pm$4.8 & 75.9$\pm$2.8 & 78.6$\pm$1.9 & 72.6$\pm$2.2 & 71.0$\pm$2.2 & 45.2$\pm$2.8 & 86.3$\pm$1.6 & 49.1$\pm$0.7 \\
            &ECC [2017] & - & - & 72.7 & 76.8 & 67.8 & - & - & - & - \\
            &DGCNN [2018] & 85.8$\pm$1.7 & 58.6$\pm$2.5 & 75.5$\pm$1.0 & 74.4$\pm$0.5 & 73.8$\pm$0.5 & 70.0$\pm$0.9 & 47.8$\pm$0.9 & 76.0$\pm$1.7 & 48.7$\pm$4.5 \\
            &\textsc{DiffPool} [2018] & 80.6 & - & 76.3 & 76.0 & 75.5 & - & - & - & - \\
            &GCAPS-CNN [2018] & - & 66.0$\pm$5.9 & 76.4$\pm$4.2 & 82.7$\pm$2.4 & 77.7$\pm$2.5 & 71.7$\pm$3.4 & 48.5$\pm$4.1 & 87.6$\pm$2.5 & 50.1$\pm$1.7 \\
            &GIN-0  + \textsc{Sum/Avg} [2018] & 89.4$\pm$5.6 & 64.6$\pm$7.0 & 76.2$\pm$2.8 & 82.7$\pm$1.7 & 80.2$\pm$1.9 & 75.1$\pm$5.1 & 52.3$\pm$2.8 & \textbf{92.4}$\pm$\textbf{2.5} & 57.5$\pm$1.5 \\
            &EigenGCN [2019] & 79.5 & - & 76.6 & 77.0 & - & - & - & - & - \\
            \midrule
            \multirow{2}{*}{\rotatebox[origin=c]{90}{Ours}}
            &GIN-0 + \textsc{SOPool}$_{attn}$ & 93.6$\pm$4.1 & 72.9$\pm$6.2 & 79.4$\pm$3.2 & 82.8$\pm$1.4 & \textbf{81.1}$\pm$\textbf{1.8} & 78.1$\pm$4.0 & 54.3$\pm$2.6 & 91.7$\pm$2.7 & 58.3$\pm$1.4 \\
            &GIN-0 + \textsc{SOPool}$_{bimap}$ & \textbf{95.3}$\pm$\textbf{4.4} & \textbf{75.0}$\pm$\textbf{4.3} & \textbf{80.1}$\pm$\textbf{2.7} & 83.6$\pm$1.4 & 79.9$\pm$1.9 & 78.4$\pm$4.7 & \textbf{54.6}$\pm$\textbf{3.6} & 89.6$\pm$3.3 & \textbf{58.4}$\pm$\textbf{1.6} \\
            &GIN-0 + \textsc{SOPool}$_{m\_attn}$ & 95.2$\pm$5.4 & 74.4$\pm$5.5 & 79.5$\pm$3.1 & \textbf{84.5}$\pm$\textbf{1.3} & 77.6$\pm$1.9 & \textbf{78.5}$\pm$\textbf{2.8} & 54.3$\pm$2.1 & 90.0$\pm$0.8 & 55.8$\pm$2.2 \\
            \bottomrule
    \end{tabular}}
\end{table*}

\begin{table*}
    \caption{Comparison results between our proposed methods and other graph pooling methods by fixing the GNN before graph pooling to GIN-0, as described in Section~\ref{sec:exp_pool}. The best models are highlighted with boldface.}
    \label{table:pool}
    \centering
    \resizebox{1.0\textwidth}{!}{
        \begin{tabular}{lccccccccc}
            \toprule
            Models & MUTAG & PTC & PROTEINS & NCI1 & COLLAB & IMDB-B & IMDB-M & RDT-B & RDT-M5K\\
            \midrule
            GIN-0 + \textsc{Sum/Avg} & 89.4$\pm$5.6 & 64.6$\pm$7.0 & 76.2$\pm$2.8 & 82.7$\pm$1.7 & 80.2$\pm$1.9 & 75.1$\pm$5.1 & 52.3$\pm$2.8 & 92.4$\pm$2.5 & 57.5$\pm$1.5 \\
            GIN-0 + \textsc{DiffPool} & 94.8$\pm$4.8 & 66.1$\pm$7.7 & 78.8$\pm$3.1 & 76.6$\pm$1.3 & 75.3$\pm$2.2 & 74.4$\pm$4.0 & 50.1$\pm$3.2 & - & - \\
            GIN-0 + \textsc{SortPool} & 95.2$\pm$3.9 & 69.5$\pm$6.3 & 79.2$\pm$3.0 & 78.9$\pm$2.7 & 78.2$\pm$1.6 & 77.5$\pm$2.7 & 53.1$\pm$2.9 & 81.6$\pm$4.6 & 48.4$\pm$4.8 \\
            GIN-0 + \textsc{TopKPool} & 94.7$\pm$3.5 & 68.4$\pm$6.4 & 79.1$\pm$2.2 & 79.6$\pm$1.7 & 79.6$\pm$2.1 & 77.8$\pm$5.1 & 53.7$\pm$2.8 & - & - \\
            GIN-0 + \textsc{SAGPool} & 93.9$\pm$3.3 & 69.0$\pm$6.6 & 78.4$\pm$3.1 & 79.0$\pm$2.8 & 78.9$\pm$1.7 & 77.8$\pm$2.9 & 53.1$\pm$2.8 & - & - \\
            \midrule
            GIN-0 + \textsc{AttnPool} & 93.2$\pm$5.8 & 71.2$\pm$8.0 & 77.5$\pm$3.3 & 80.6$\pm$2.1 & \textbf{81.8}$\pm$\textbf{2.2} & 77.1$\pm$4.4 & 53.8$\pm$2.5 & \textbf{92.5}$\pm$\textbf{2.3} & 57.9$\pm$1.7 \\
            GIN-0 + \textsc{SOPool}$_{attn}$ & 93.6$\pm$4.1 & 72.9$\pm$6.2 & 79.4$\pm$3.2 & 82.8$\pm$1.4 & 81.1$\pm$1.8 & 78.1$\pm$4.0 & 54.3$\pm$2.6 & 91.7$\pm$2.7 & 58.3$\pm$1.4 \\
            \midrule
            GIN-0 + \textsc{CovPool} & \textbf{95.3}$\pm$\textbf{3.7} & 73.3$\pm$5.1 & \textbf{80.1}$\pm$\textbf{2.2} & 83.5$\pm$1.9 & 79.3$\pm$1.8 & 72.1$\pm$5.1 & 47.8$\pm$2.7 & 90.3$\pm$3.6 & \textbf{58.4}$\pm$\textbf{1.7} \\
            GIN-0 + \textsc{SOPool}$_{bimap}$ & \textbf{95.3}$\pm$\textbf{4.4} & \textbf{75.0}$\pm$\textbf{4.3} & \textbf{80.1}$\pm$\textbf{2.7} & 83.6$\pm$1.4 & 79.9$\pm$1.9 & 78.4$\pm$4.7 & \textbf{54.6}$\pm$\textbf{3.6} & 89.6$\pm$3.3 & \textbf{58.4}$\pm$\textbf{1.6} \\
            \midrule
            GIN-0 + \textsc{SOPool}$_{m\_attn}$ & 95.2$\pm$5.4 & 74.4$\pm$5.5 & 79.5$\pm$3.1 & \textbf{84.5}$\pm$\textbf{1.3} & 77.6$\pm$1.9 & \textbf{78.5}$\pm$\textbf{2.8} & 54.3$\pm$2.1 & 90.0$\pm$0.8 & 55.8$\pm$2.2 \\
            \bottomrule
    \end{tabular}}
\end{table*}

More statistics of these datasets are provided in the ``datasets'' section of Table~\ref{table:baseline}. The input node features are different for different datasets. For bioinformatics datasets, the nodes have categorical labels as input features. For social network datasets, we create node features. To be specific, we set all node feature vectors to be the same for REDDIT-BINARY and REDDIT-MULTI5K~\cite{xu2019powerful}. And for the other social network datasets, we use one-hot encoding of node degrees as features.

\textbf{Configurations.} In Sections \ref{sec:exp_baseline}, \ref{sec:exp_pool} and \ref{sec:exp_encoder}, the flat GNNs we use with our proposed graph pooling methods are graph isomorphism networks~(GINs)~\cite{xu2019powerful}. The original GINs employ averaging or summation~(\textsc{Sum/Avg}) as the graph pooling function; specifically, summation on bioinformatics datasets and averaging on social datasets. We replace averaging or summation with our proposed graph pooling methods and keep other parts the same. There are seven variants of GINs, two of which are equivalent to graph convolutional network~(GCN)~\cite{kipf2017semi} and GraphSAGE~\cite{hamilton2017inductive}, respectively. In Sections~\ref{sec:exp_baseline} and~\ref{sec:exp_pool}, we use GIN-0 with our methods. In Section~\ref{sec:exp_encoder}, we examine our methods with all variants of GINs. Details of all variants can be found in Section~\ref{sec:exp_encoder}.

The hierarchical GNNs used in Section~\ref{sec:exp_hierarchical} follow the hierarchical architecture in~\cite{lee2019self}, allowing direct comparisons. To be specific, each block is composed of one GNN layer followed by a hierarchical graph pooling. After each hierarchical pooling, a classifier is used. The final prediction is the combination of all classifiers.

\textbf{Training \& Evaluation.} Following~\cite{yanardag2015deep,niepert2016learning}, model performance is evaluated using 10-fold cross-validation and reported as the average and standard deviation of validation accuracies across the 10 folds. For the flat GNNs, we follow the same training process in~\cite{xu2019powerful}. All GINs have 5 layers. Each multi-layer perceptron~(MLP) has 2 layers with batch normalization~\cite{ioffe2015batch}. For the hierarchical GNNs, we follow the the same training process in~\cite{lee2019self}. There are three blocks in total. Dropout~\cite{srivastava2014dropout} is applied in the classifiers. The Adam optimizer~\cite{kingma2014adam} is used with the learning rate initialized as 0.01 and decayed by 0.5 every 50 epochs. The number of total epochs is selected according to the best cross-validation accuracy. We tune the number of hidden units (16, 32, 64) and the batch size (32, 128) using grid search. 

\textbf{Baselines.} We compare our methods with various graph classification models as baselines, including both kernel-based and GNN-based methods. The kernel-based methods are graphlet kernel~(GK)~\cite{shervashidze2009efficient}, random walk kernel~(RW)~\cite{vishwanathan2010graph}, Weisfeiler-Lehman subtree kernel~(WL)~\cite{shervashidze2011weisfeiler}, deep graphlet kernel~(DGK)~\cite{yanardag2015deep}, and anonymous walk embeddings~(AWE)~\cite{ivanov2018anonymous}. Among them, DGK and AWE use deep learning methods as well. The GNN-based methods are diffusion-convolutional neural network~(DCNN)~\cite{atwood2016diffusion}, \textsc{PatchScan}~\cite{niepert2016learning}, ECC~\cite{simonovsky2017dynamic}, deep graph CNN~(DGCNN)~\cite{zhang2018end}, differentiable pooling~(\textsc{DiffPool})~\cite{ying2018hierarchical}, graph capsule CNN~(GCAPS-CNN)~\cite{verma2018graph}, self-attention graph pooling~(\textsc{SAGPool})~\cite{lee2019self}, GIN~\cite{xu2019powerful}, and eigenvector-based pooling~(EigenGCN)~\cite{Ma:2019:GCN:3292500.3330982}. We report the performance of these baselines provided in \cite{zhang2018end,verma2018graph,xu2019powerful,ying2018hierarchical,lee2019self,Ma:2019:GCN:3292500.3330982}.

\subsection{Comparison with Baselines}\label{sec:exp_baseline}

\begin{table*}
    \caption{Results of our proposed methods with different GNNs before graph pooling, as described in Section~\ref{sec:exp_encoder}. The architectures of different GNNs come from variants of GINs in~\cite{xu2019powerful}, whose details can be found in the supplementary material. The best models are highlighted with boldface.}
    \label{table:gnns}
    \centering
    \resizebox{1.0\textwidth}{!}{
        \begin{tabular}{llccccccc}
            \toprule
            GNNs & \textsc{Pool}s & MUTAG & PTC & PROTEINS & NCI1 & COLLAB & IMDB-B & IMDB-M \\
            \midrule
            \multirow{3}{*}{\textsc{Sum-MLP} (GIN-0)}
            &\textsc{Sum/Avg} & 89.4$\pm$5.6 & 64.6$\pm$7.0 & 76.2$\pm$2.8 & 82.7$\pm$1.7 & 80.2$\pm$1.9 & 75.1$\pm$5.1 & 52.3$\pm$2.8 \\
            &\textsc{SOPool}$_{attn}$ & 93.6$\pm$4.1 & 72.9$\pm$6.2 & 79.4$\pm$3.2 & 82.8$\pm$1.4 & \textbf{81.1}$\pm$\textbf{1.8} & 78.1$\pm$4.0 & 54.3$\pm$2.6 \\
            &\textsc{SOPool}$_{bimap}$ & \textbf{95.3}$\pm$\textbf{4.4} & \textbf{75.0}$\pm$\textbf{4.3} & \textbf{80.1}$\pm$\textbf{2.7} & \textbf{83.6}$\pm$\textbf{1.4} & 79.9$\pm$1.9 & \textbf{78.4}$\pm$\textbf{4.7} & \textbf{54.6}$\pm$\textbf{3.6} \\
            \midrule
            \multirow{3}{*}{\textsc{Sum-MLP} (GIN-$\epsilon$)}
            &\textsc{Sum/Avg} & 89.0$\pm$6.0 & 63.7$\pm$8.2 & 75.9$\pm$3.8 & 82.7$\pm$1.6 & 80.1$\pm$1.9 & 74.3$\pm$5.1 & 52.1$\pm$3.6 \\
            &\textsc{SOPool}$_{attn}$ & 92.6$\pm$5.4 & \textbf{73.6}$\pm$\textbf{5.5} & 79.2$\pm$1.9 & 83.1$\pm$1.8 & \textbf{80.6}$\pm$\textbf{1.6} & \textbf{78.1}$\pm$\textbf{4.3} & \textbf{55.4}$\pm$\textbf{3.7} \\
            &\textsc{SOPool}$_{bimap}$ & \textbf{93.7}$\pm$\textbf{5.3} & 73.5$\pm$7.0 & \textbf{79.3}$\pm$\textbf{1.8} & \textbf{83.6}$\pm$\textbf{1.4} & 80.4$\pm$2.4 & 77.5$\pm$4.5 & 54.5$\pm$3.5 \\
            \midrule
            \multirow{3}{*}{\textsc{Sum-1-Layer}}
            &\textsc{Sum/Avg} & 90.0$\pm$8.8 & 63.1$\pm$5.7 & 76.2$\pm$2.6 & 82.0$\pm$1.5 & 80.6$\pm$1.9 & 74.1$\pm$5.0 & 52.2$\pm$2.4 \\
            &\textsc{SOPool}$_{attn}$ & 94.2$\pm$4.4 & \textbf{73.6}$\pm$\textbf{6.5} & 79.0$\pm$2.9 & 81.2$\pm$1.5 & \textbf{81.2}$\pm$\textbf{1.6} & \textbf{78.6}$\pm$\textbf{4.1} & \textbf{54.5}$\pm$\textbf{3.0} \\
            &\textsc{SOPool}$_{bimap}$ & \textbf{95.8}$\pm$\textbf{4.2} & 71.8$\pm$6.1 & \textbf{80.1}$\pm$\textbf{2.5} & \textbf{82.4}$\pm$\textbf{1.3} & 80.5$\pm$2.0 & 78.2$\pm$3.6 & 54.1$\pm$3.4 \\
            \midrule
            \multirow{3}{*}{\textsc{Mean-MLP}}
            &\textsc{Sum/Avg} & 83.5$\pm$6.3 & 66.6$\pm$6.9 & 75.5$\pm$3.4 & 80.9$\pm$1.8 & 79.2$\pm$2.3 & 73.7$\pm$3.7 & 52.3$\pm$3.1 \\
            &\textsc{SOPool}$_{attn}$ & \textbf{92.6}$\pm$\textbf{4.5} & \textbf{74.9}$\pm$\textbf{6.6} & \textbf{79.4}$\pm$\textbf{2.8} & 80.6$\pm$1.1 & 80.0$\pm$2.0 & 77.5$\pm$3.9 & \textbf{55.2}$\pm$\textbf{3.3} \\
            &\textsc{SOPool}$_{bimap}$ & 90.4$\pm$6.2 & 72.7$\pm$4.0 & 79.3$\pm$2.4 & \textbf{81.1}$\pm$\textbf{1.6} & \textbf{80.4}$\pm$\textbf{1.7} & \textbf{77.9}$\pm$\textbf{4.7} & 55.0$\pm$3.7 \\
            \midrule
            \multirow{3}{*}{\textsc{Mean-1-Layer} (GCN)}
            &\textsc{Sum/Avg} & 85.6$\pm$5.8 & 64.2$\pm$4.3 & 76.0$\pm$3.2 & \textbf{80.2}$\pm$\textbf{2.0} & 79.0$\pm$1.8 & 74.0$\pm$3.4 & 51.9$\pm$3.8 \\
            &\textsc{SOPool}$_{attn}$ & 90.0$\pm$5.1 & \textbf{76.7}$\pm$\textbf{5.6} & 78.5$\pm$2.8 & 78.0$\pm$1.8 & 80.2$\pm$1.6 & \textbf{78.9}$\pm$\textbf{4.2} & \textbf{54.8}$\pm$\textbf{3.1} \\
            &\textsc{SOPool}$_{bimap}$ & \textbf{90.9}$\pm$\textbf{5.7} & 70.9$\pm$4.1 & \textbf{78.7}$\pm$\textbf{3.1} & 78.8$\pm$1.1 & \textbf{80.4}$\pm$\textbf{2.1} & 77.7$\pm$4.5 & 54.5$\pm$4.0 \\
            \midrule
            \multirow{3}{*}{\textsc{Max-MLP}}
            &\textsc{Sum/Avg} & 84.0$\pm$6.1 & 64.6$\pm$10.2 & 76.0$\pm$3.2 & 77.8$\pm$1.3 & - & 73.2$\pm$5.8 & 51.1$\pm$3.6 \\
            &\textsc{SOPool}$_{attn}$ & \textbf{90.0}$\pm$\textbf{7.3} & 72.4$\pm$4.7 & 78.3$\pm$3.1 & \textbf{78.6}$\pm$\textbf{1.9} & - & 78.1$\pm$4.1 & 54.1$\pm$3.4 \\
            &\textsc{SOPool}$_{bimap}$ & 88.8$\pm$7.0 & \textbf{73.3}$\pm$\textbf{5.5} & \textbf{78.4}$\pm$\textbf{3.0} & 78.0$\pm$1.9 & - & \textbf{78.2}$\pm$\textbf{4.7} & \textbf{54.6}$\pm$\textbf{3.5} \\
            \midrule
            \multirow{3}{*}{\textsc{Max-1-Layer} (GraphSAGE)}
            &\textsc{Sum/Avg} & 85.1$\pm$7.6 & 63.9$\pm$7.7 & 75.9$\pm$3.2 & \textbf{77.7}$\pm$\textbf{1.5} & - & 72.3$\pm$5.3 & 50.9$\pm$2.2 \\
            &\textsc{SOPool}$_{attn}$ & \textbf{90.0}$\pm$\textbf{6.8} & 72.1$\pm$5.9 & \textbf{79.0}$\pm$\textbf{2.9} & 77.4$\pm$1.8 & - & 77.4$\pm$5.1 & 54.1$\pm$3.1 \\
            &\textsc{SOPool}$_{bimap}$ & 89.9$\pm$5.8 & \textbf{73.6}$\pm$\textbf{5.1} & 78.9$\pm$2.8 & 77.0$\pm$2.0 & - & \textbf{78.6}$\pm$\textbf{4.7} & \textbf{54.2}$\pm$\textbf{3.9} \\
            \bottomrule
    \end{tabular}}
\end{table*}

The comparison results between our methods and baselines are reported in Table~\ref{table:baseline}. GIN-0 equipped with our proposed graph pooling methods, ``GIN-0 + \textsc{SOPool}$_{attn}$'', ``GIN-0 + \textsc{SOPool}$_{bimap}$'', and ``GIN-0 + \textsc{SOPool}$_{m\_attn}$'', outperform all the baselines significantly on seven out of nine datasets. On NCI1, WL has better performance than all GNN-based models. However, ``GIN-0 + \textsc{SOPool}$_{m\_attn}$'' is the second best model and has improved performance over other GNN-based models. On REDDIT-BINARY, our methods achieve comparable performance to the best one.

It is worth noting that the baseline ``GIN-0  + \textsc{Sum/Avg}'' is the previous state-of-the-art model~\cite{xu2019powerful}. Our methods differ from it only in the graph pooling functions. The significant improvement demonstrates the effectiveness of our proposed graph pooling methods. In the next section, we compare our methods with other graph pooling methods by fixing the GNN before graph pooling to GIN-0, in order to eliminate the influences of different GNNs.

\subsection{Ablation Studies in Flat Graph Neural Networks}\label{sec:exp_pool}

We perform ablation studies to show that our proposed methods are superior to other global graph pooling methods under a fair setting. Starting from the baseline ``GIN-0  + \textsc{Sum/Avg}'', we replace \textsc{Sum/Avg} with different graph pooling methods and keep all other configurations unchanged. The graph pooling methods we include are \textsc{DiffPool}~\cite{ying2018hierarchical}, \textsc{SortPool} from DGCNN~\cite{zhang2018end}, \textsc{TopKPool} from Graph U-Net~\cite{gao2018graph}, \textsc{SAGPool}~\cite{lee2019self}, and $\textsc{CovPool}$ and $\textsc{AttnPool}$ described in Section~\ref{sec:method_discuss}. \textsc{DiffPool}, \textsc{TopKPool}, and \textsc{SAGPool} are used as hierarchical graph pooling methods in their works, but they achieve good performance as global pooling methods as well~\cite{ying2018hierarchical,lee2019self}. \textsc{EigenPool} from EigenGCN suffers from significant performance loss as a global pooling method~\cite{Ma:2019:GCN:3292500.3330982} so that we do not include it in the ablation studies. $\textsc{CovPool}$ and $\textsc{AttnPool}$ use the same settings as our proposed methods.

Table~\ref{table:pool} provides the comparison results. Our proposed \textsc{SOPool}$_{bimap}$ and \textsc{SOPool}$_{attn}$ achieve better performance than \textsc{DiffPool}, \textsc{SortPool}, \textsc{TopKPool}, and \textsc{SAGPool} on all datasets, demonstrating the effectiveness of our graph pooling methods with second-order statistics.

To support our discussion in Section~\ref{sec:method_discuss}, we analyze the performance of $\textsc{CovPool}$ and $\textsc{AttnPool}$. Note that the same bilinear mapping technique used in \textsc{SOPool}$_{bimap}$ is applied on $\textsc{CovPool}$, in order to avoid the excessive number of parameters. $\textsc{CovPool}$ achieves comparable performance to \textsc{SOPool}$_{bimap}$ on most datasets. However, huge performance loss is observed on PTC, IMDB-BINARY, and IMDB-MULTI, indicating that subtracting the mean is harmful in graph pooling.

Compared to \textsc{SOPool}$_{attn}$, $\textsc{AttnPool}$ suffers from performance loss on all datasets except COLLAB and REDDIT-BINARY. The loss is especially significant on bioinformatics datasets (PTC, PROTEINS, NCI1). However, $\textsc{AttnPool}$ achieves the best performance on COLLAB and REDDIT-BINARY among all graph pooling methods, although the added \textsc{Softmax} function results in less discriminative power. The reason might be capturing the distributional information is more important than the exact structure in these datasets. It is similar to GINs, where using averaging as graph pooling achieves better performance on social network datasets than summation~\cite{xu2019powerful}.

\subsection{Results with Different Graph Neural Networks}\label{sec:exp_encoder}

We've already demonstrated the superiority of our proposed \textsc{SOPool}$_{bimap}$ and \textsc{SOPool}$_{attn}$ over previous pooling methods. Next, we show that their effectiveness is robust to different GNNs. In this experiment, we change GIN-0 into other six variants of GINs. Note that these variants cover Graph Convolutional Networks (GCN)~\cite{kipf2017semi} and GraphSAGE~\cite{hamilton2017inductive}, thus including a wide range of different kinds of GNNs.

We first give details of different variants of graph isomorphism networks~(GINs)~\cite{xu2019powerful}. Basically, GINs iteratively update the representation of each node in a graph by aggregating representations of its neighbors, where the iteration is achieved by stacking several layers. Therefore, it suffice to describe the $k$-th layer of GINs based on one node.

Recall that we represent a graph $G = (A, X)$ by its adjacency matrix $A \in \{0,1\}^{n \times n}$ and node feature matrix $X \in \mathbb{R}^{n \times d}$, where $n$ is the number of nodes in $G$ and $d$ is the dimension of node features. The adjacency matrix tells the neighboring information of each node. We introduce GINs by defining node representation matrices $H^{(k-1)} \in \mathbb{R}^{n \times f^{(k-1)}}$ and $H^{(k)} \in \mathbb{R}^{n \times f^{(k)}}$ as inputs and outputs to the $k$-th layer, respectively. We have $H^{(0)} = X$. Note that the first dimension $n$ does not change during the computation, as GINs learn representations for each node.

Specifically, consider a node $\nu$ has corresponding representations $h_\nu^{(k-1)} \in \mathbb{R}^{f^{(k-1)}}$ and $h_\nu^{(k)} \in \mathbb{R}^{f^{(k)}}$, which are rows of $H^{(k-1)}$ and $H^{(k)}$, respectively. The set of neighboring nodes of $\nu$ is given by $\mathcal{N}(\nu)$. We describe the $k$-layer of the following variants:

\begin{itemize}
    \item \textsc{\textbf{Sum-MLP (GIN-0):}} $$h_\nu^{(k)} = \text{MLP}^{(k)}(h_\nu^{(k-1)} + \sum_{\mu \in \mathcal{N}(\nu)} h_\mu^{(k-1)})$$
    \item \textsc{\textbf{Sum-MLP (GIN-$\epsilon$):}} $$h_\nu^{(k)} = \text{MLP}^{(k)}((1+\epsilon^{(k)})h_\nu^{(k-1)} + \sum_{\mu \in \mathcal{N}(\nu)} h_\mu^{(k-1)})$$
    \item \textsc{\textbf{Sum-1-Layer:}} $$h_\nu^{(k)} = \text{ReLU}(W^{(k)}(h_\nu^{(k-1)} + \sum_{\mu \in \mathcal{N}(\nu)} h_\mu^{(k-1)}))$$
    \item \textsc{\textbf{Mean-MLP:}} $$h_\nu^{(k)} = \text{MLP}^{(k)}(\textsc{Mean}\{h_\mu^{(k-1)}, \forall \mu \in \nu \cup \mathcal{N}(\nu)\})$$
    \item \textsc{\textbf{Mean-1-Layer (GCN):}} $$h_\nu^{(k)} = \text{ReLU}(W^{(k)}(\textsc{Mean}\{h_\mu^{(k-1)}, \forall \mu \in \nu \cup \mathcal{N}(\nu)\})$$
    \item \textsc{\textbf{Max-MLP:}} $$h_\nu^{(k)} = \text{MLP}^{(k)}(\textsc{Max}\{h_\mu^{(k-1)}, \forall \mu \in \nu \cup \mathcal{N}(\nu)\})$$
    \item \textbf{\textsc{Max-1-Layer} (GraphSAGE):} $$h_\nu^{(k)} = \text{ReLU}(W^{(k)}(\textsc{Max}\{h_\mu^{(k-1)}, \forall \mu \in \nu \cup \mathcal{N}(\nu)\})$$
\end{itemize}

Here, the multi-layer perceptron~(MLP) has two layers with ReLU activation functions. Note that \textsc{Mean-1-Layer} and \textsc{Max-1-Layer} correspond to GCN~\cite{kipf2017semi} and GraphSAGE~\cite{hamilton2017inductive}, respectively, up to minor architecture modifications.

The results of these different GNNs with our graph pooling methods are reported in Table~\ref{table:gnns}. Our proposed \textsc{SOPool}$_{bimap}$ and \textsc{SOPool}$_{attn}$ achieve satisfying performance consistently. In particular, on social network datasets, the performance does not decline when the GNNs before graph pooling become less powerful, showing the highly discriminative ability of second-order pooling.

\subsection{Ablation Studies in Hierarchical Graph Neural Networks}\label{sec:exp_hierarchical}

\begin{table}
	\caption{Comparison results between different hierarchical graph pooling methods. The hierarchical GNN architecture follows the one in~\cite{lee2019self}. We report the accuracies of the baselines provided in \cite{lee2019self}. The best models are highlighted with boldface.}
	\label{table:hpool}
	\centering
	\begin{tabular}{lcc}
		\toprule
		Models & DD & PROTEINS \\
		\midrule
		\textsc{DiffPool} & 67.0$\pm$2.4 & 68.2$\pm$2.0 \\
		\textsc{TopKPool} & 75.0$\pm$0.9 & 71.1$\pm$0.9 \\
		\textsc{SAGPool} & 76.5$\pm$1.0 & 71.9$\pm$1.0 \\
		\midrule
		\textsc{SOPool}$_{m\_attn}$ & \textbf{76.8}$\pm$\textbf{1.9} & \textbf{77.1}$\pm$\textbf{3.8} \\
		\bottomrule
	\end{tabular}
\end{table}

\begin{table}
	\caption{Comparison results of \textsc{SOPool}$_{m\_attn}$ with different hierarchical GNNs. The hierarchical GNN architecture follows the one in~\cite{lee2019self}, where we change the number of blocks from one to three. The best models are highlighted with boldface.}
	\label{table:hpool_n}
	\centering
	\begin{tabular}{lcc}
		\toprule
		Models & DD & PROTEINS \\
		\midrule
		1 block & 73.3$\pm$2.4 & 77.4$\pm$4.3 \\
		2 blocks & \textbf{77.2}$\pm$\textbf{2.7} & \textbf{78.1}$\pm$\textbf{4.3} \\
		3 blocks & 76.8$\pm$1.9 & 77.1$\pm$3.8 \\
		\bottomrule
	\end{tabular}
\end{table}

\textsc{SOPool}$_{m\_attn}$ has shown its effectiveness as global graph pooling through the experiments in Sections~\ref{sec:exp_baseline} and \ref{sec:exp_pool}. In this section, we evaluate it as hierarchical graph pooling in hierarchical GNNs. The hierarchical GNN architecture follows the one in~\cite{lee2019self}, which contains three blocks of a GNN layer followed by graph pooling, as introduced in Section~\ref{sec:exp_setup}. The experiments are performed on DD and PROTEINS datasets, where hierarchical GNNs tend to achieve good performance~\cite{ying2018hierarchical,lee2019self}.

First, we compare \textsc{SOPool}$_{m\_attn}$ with different hierarchical graph pooling methods under the same hierarchical GNN architecture. Specifically, we include \textsc{DiffPool}, \textsc{TopKPool}, and \textsc{SAGPool}, which have been used as hierarchical graph pooling methods in their works. The comparison results are provided in Table~\ref{table:hpool}. Our proposed \textsc{SOPool}$_{m\_attn}$ outperforms all the baselines on both datasets, indicating the effectiveness of \textsc{SOPool}$_{m\_attn}$ as a hierarchical graph pooling method.

In addition, we conduct experiments to evaluate \textsc{SOPool}$_{m\_attn}$ in different hierarchical GNNs by varying the number of blocks. The results are shown in Table~\ref{table:hpool_n}. On both datasets, \textsc{SOPool}$_{m\_attn}$ achieves the best performance when the number of blocks is two. The results indicate current datasets on graph classification are not large enough yet. And without techniques like jumping knowledge networks~(JK-Net)~\cite{xu2018representation}, hierarchical GNNs tend to suffer from over-fitting, leading to worse performance than flat GNNs.

\section{Conclusions}\label{sec:conclu}
In this work, we propose to perform graph representation learning with second-order pooling, by pointing out that second-order pooling can naturally solve the challenges of graph pooling. Second-order pooling is more powerful than existing graph pooling methods, since it is capable of using all node information and collecting second-order statistics that encode feature correlations and topology information. To take advantage of second-order pooling in graph representation learning, we propose two global graph pooling approaches based on second-order pooling; namely, bilinear mapping and attentional second-order pooling. Our proposed methods solve the practical problems incurred by directly using second-order pooling with GNNs. We theoretically show that our proposed methods are more suitable to graph representation learning by comparing with two related pooling methods from computer vision tasks. In addition, we extend one of the proposed method to a hierarchical graph pooling method, which has more flexibility. To demonstrate the effectiveness of our methods, we conduct thorough experiments on graph classification tasks. Our proposed methods have achieved the new state-of-the-art performance on eight out of nine benchmark datasets. Ablation studies are performed to show that our methods outperform existing graph pooling methods significantly and achieve good performance consistently with different GNNs.

\ifCLASSOPTIONcompsoc
  \section*{Acknowledgments}
\else
  \section*{Acknowledgment}
\fi

This work was supported in part by National Science Foundation grant
IIS-1908166, and Defense Advanced Research Projects Agency grant
N66001-17-2-4031.

\ifCLASSOPTIONcaptionsoff
  \newpage
\fi



\bibliographystyle{IEEEtran}
\bibliography{reference}
\end{document}